\documentclass[a4paper, 10pt, conference]{ieeeconf}

\let\labelindent\relax
\usepackage{graphics}
\usepackage{epsfig}
\usepackage{mathptmx}
\usepackage{times}
\usepackage{amsmath}
\usepackage{amssymb}
\usepackage{xcolor}
\usepackage{subfigure}
\usepackage{graphicx}
\usepackage{caption}
\usepackage{enumitem}
\usepackage{hyperref}
\hypersetup{
    colorlinks=true,
    linkcolor=blue,
    filecolor=black,
    urlcolor=blue,
    citecolor=blue,
}

\usepackage[backend=biber]{biblatex}
\newcommand{\ie}{\textit{i}.\textit{e}. }
\newcommand{\eg}{\textit{e}.\textit{g}. }
\newcommand\sref{Section~\ref}

\title{\LARGE \bf
A Practical Approach to Insertion with Variable Socket Position \\
Using Deep Reinforcement Learning
}

\author{Mel Vecerik, Oleg Sushkov, David Barker, Thomas Roth{\"o}rl, Todd Hester, Jon Scholz$^{1}$%
\thanks{$^{1}$DeepMind, London, UK
        {\tt\small Correspondence to: \{vec,jscholz\}@google.com}}%
}

\begin{document}

\maketitle
\thispagestyle{empty}
\pagestyle{empty}

\begin{abstract}

Insertion is a challenging haptic and visual control problem with significant practical value for manufacturing.
Existing approaches in the model-based robotics community can be highly effective when task geometry is known, but are complex and cumbersome to implement, and must be tailored to each individual problem by a qualified engineer.
Within the learning community there is a long history of insertion research, but existing approaches are typically either too sample-inefficient to run on real robots, or assume access to high-level object features, e.g. socket pose.
In this paper we show that relatively minor modifications to an off-the-shelf Deep-RL algorithm (DDPG), combined with a small number of human demonstrations, allows the robot to quickly learn to solve these tasks efficiently and robustly.
Our approach requires no modeling or simulation, no parameterized search or alignment behaviors, no vision system aside from raw images, and no reward shaping.
We evaluate our approach on a narrow-clearance peg-insertion task and a deformable clip-insertion task, both of which include variability in the socket position.
Our results show that these tasks can be solved reliably on the real robot in less than 10 minutes of interaction time, and that the resulting policies are robust to variance in the socket position and orientation.

\end{abstract}

\section{INTRODUCTION}
\label{sec:introduction}
Object insertion is a long-standing benchmark task in robotics with both theoretical and practical value.
For the research community, it challenges our algorithms to cope with subtle aspects of object geometry and contact, and how to predict these quantities (perhaps implicitly) from the object's appearance.
For the wider automation community, insertion is a common use case for purchasing a robot, and a large number of industrial robotics applications involve some form of insertion.
In the vast majority of existing work on insertion the robot is blind: the socket-pose is assumed to be constant, and goals are either defined using known socket geometry, or via the robot's kinematics.
Even in this well-structured setting the peg can easily become jammed in narrow-clearance cases when using standard force controllers.  This issue has motivated various attempts to create recovery behaviors, but to-date the classical peg-in-hole problem is still considered unsolved \cite{kronander2014task}.

In real-world settings the insertion task can be considerably harder due to uncertainty in the socket or robot position, and non-rigidity in the grasp configuration or the object itself (\eg wire, soft-plastic).
With the rise of collaborative robots, there is also increasing pressure to create planning and control algorithms which can be configured by end-users with little or no formal engineering training.
Therefore the ideal solution would be robust to position and dynamics uncertainty, and not require modeling, simulation, object trackers, reward shaping, or manual tuning of control parameters.
In this paper we present an Efficient Reinforcement learning Insertion Approach based on Demonstrations, EDRIAD.
Our long-term vision is for end-users to be able to teach a robot to solve complex insertion tasks in natural settings with only a few examples of the task being solved, and for the robot quickly \textit{generalize} and \textit{improve performance} up to the limits of the physical system.

\begin{figure}[!tbp]
  \centering
  \begin{minipage}[b]{0.23\textwidth}
    \includegraphics[width=\textwidth]{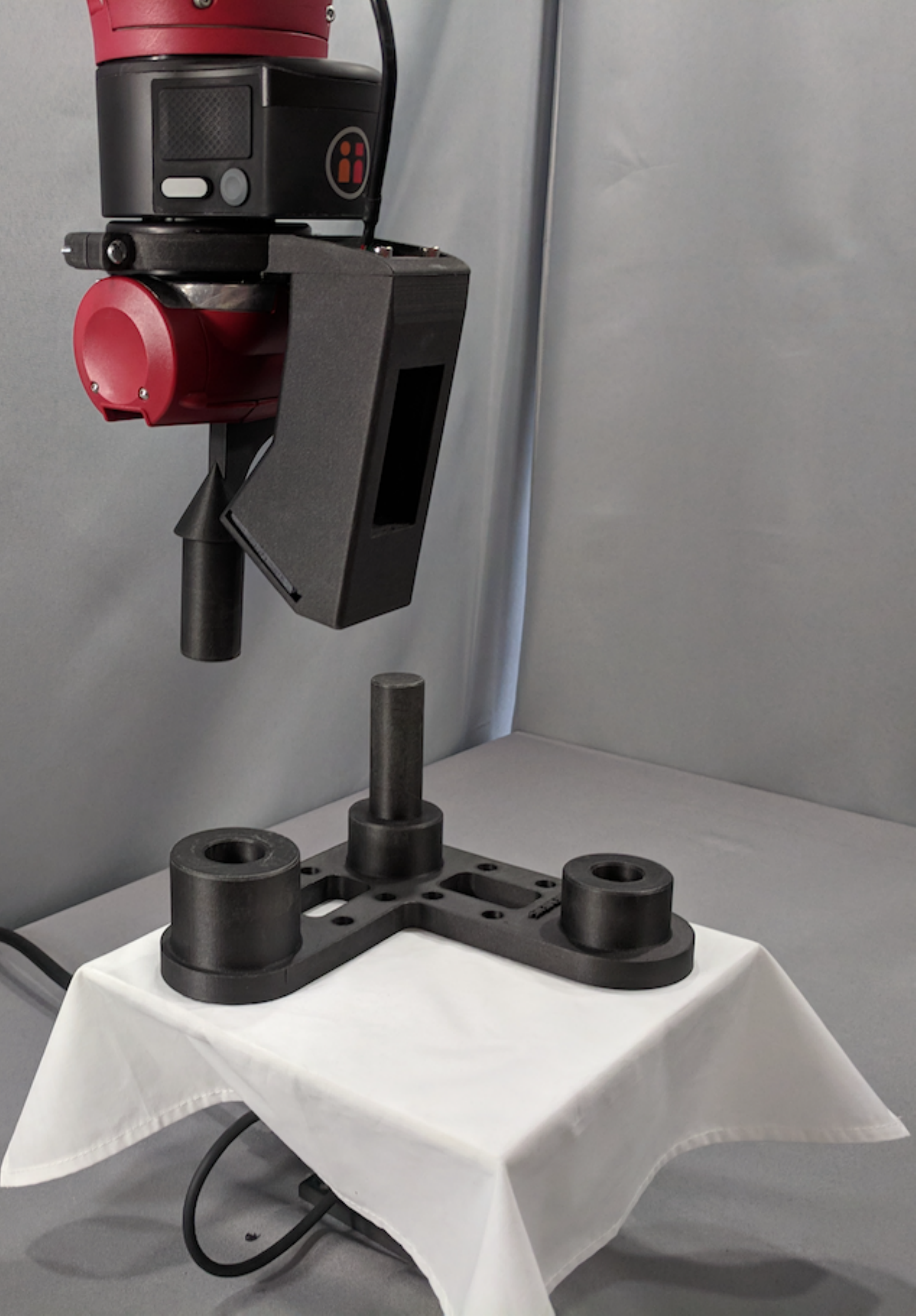}
    \caption{Peg insertion task ($0.5$ mm clearance).}
  \end{minipage}
  \hfill
  \begin{minipage}[b]{0.23\textwidth}
    \includegraphics[width=\textwidth]{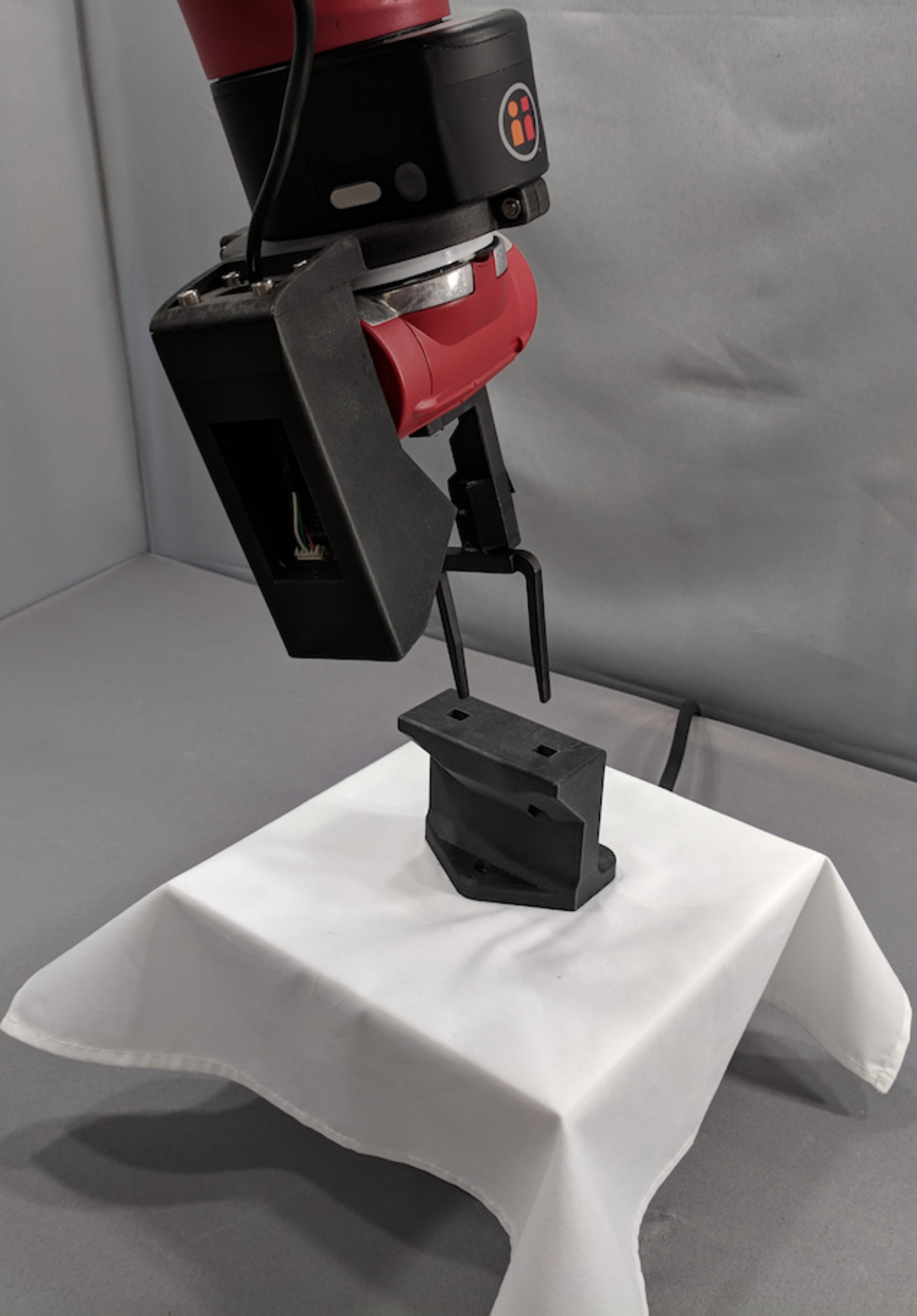}
    \caption{Clip insertion task (deformable).}
  \end{minipage}
\end{figure}

Recent work \cite{hester2017learning,matas2018sim} has shown that off-policy RL can be combined with human demonstrations to allow neural-network controllers to be trained efficiently, even when using a sparse reward function (\eg a goal classifier).  \cite{vecerik2017leveraging} extended this approach to the continuous-actions regime using the \textit{deep deterministic policy gradient} algorithm \cite{lillicrap2015continuous}, and showed that the resulting algorithm (DDPGfD) is capable of solving a deformable-object insertion task.
However these works did not consider stochasticity in the environment (\eg socket pose), or provide the rigorous evaluation on the real-robot that RL algorithms typically see in simulation.
For such work to be useful to a wider audience the results must be reproducible and robust.
Therefore our goal in this paper is to detail the steps required to replicate our results, including both the experiment setup and the algorithm itself.

The primary contributions of this paper are as follows:
\begin{enumerate}
    \item We incorporate vision into both the task definition (\ie visual rewards) and the robot's policy network.
    \item We add a behavior-cloning loss on the policy, and \textit{demonstration-classification} and \textit{episode-progress} losses on critic network to extract additional information from the demonstrations.
    \item We replace the traditional TD-loss on the critic network with a distributional loss \cite{bellemare2017distributional}.
    \item We perform rigorous evaluation of the performance on the real-robot across two insertion tasks with randomized target positions.
\end{enumerate}

Overall these improvements reduce the training time by roughly 60x vs. \cite{vecerik2017leveraging}, while also introducing socket randomization.
Our approach can solve our most challenging task in under $20$ minutes of robot interaction-time\footnote{Overall training time includes environment-reset and synchronous network training, neither of which we optimized for.}, which is approaching levels feasible for deployment in practical settings.
To our knowledge the only other generic end-user programmable method capable of this sample-efficiency is Guided Policy Search, which has some limitations discussed in \sref{sec:related_work}.

Similar to \cite{mahmood2018setting},  we also note that a considerable portion of the effort in getting a deep-RL algorithm to work on a robot was on the environment-side.
Our solution shared many aspects with \cite{mahmood2018setting}, but focused on insertion rather than reaching, and required several additional components to handle contact safely.
\sref{sec:appendix} discusses our overall experiment setup, in the hope of making our results fully reproducible in another lab or manufacturing setting.

\section{TASK DESCRIPTION}
\label{sec:task_description}

\begin{figure}[t]
 \centering
 \includegraphics[scale=0.15]{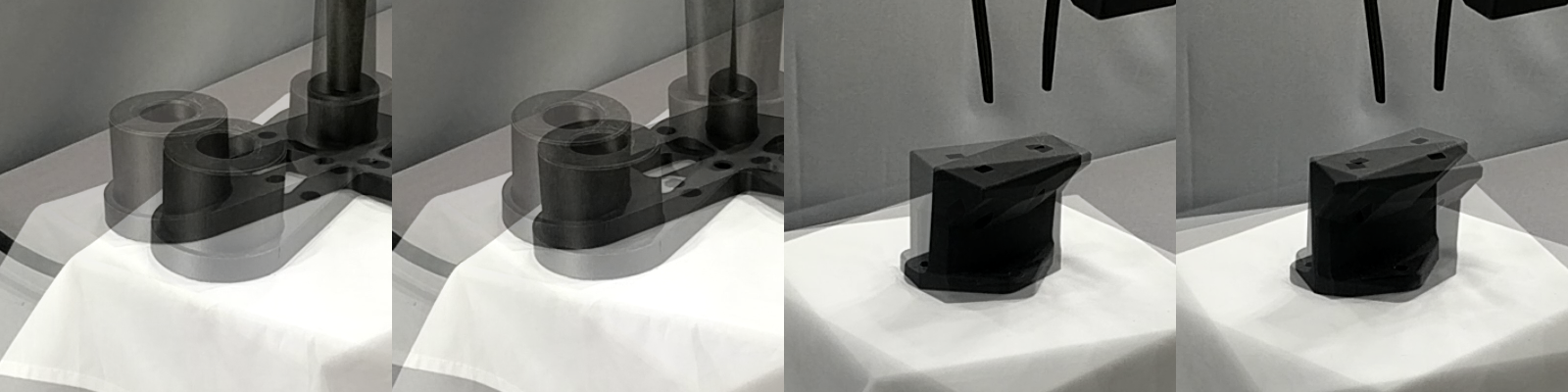}
 \caption{
  Illustration of socket pose variability.
  Each image is a blend of two frames created at the minimum and maximum of one of the degrees-of-freedom of the pan-tilt (\ie pan or tilt).
  At the beginning of each episode we sample from a Uniform distribution with these ranges in order to force the agent to be robust to socket-pose noise.}
  \label{fig:randomness_imgs}
\end{figure}

Our primary objective in this paper is to develop an algorithm that is directly applicable to real-world insertion tasks.
Therefore the task setting is of central importance in dictating our overall approach.
Based on \cite{inoue2017deep,gullapalli1994learning} we see the value of RL specifically in the final contact-rich phase of insertion tasks which are notoriously difficult to model.
While it would be convenient to learn a single monolithic policy to control the robot from arbitrary starting configurations, current Deep-RL methods lack the efficiency or safety guarantees to make this practical, and free-space motion is well handled by established motion-planning techniques \cite{latombe2012robot}.
This ``last-inch'' task setting assumes that the end effector is \textit{physically close} to the target, but otherwise makes no assumptions about the scene geometry or dynamics.
Furthermore we require that both the state and goal/reward can be defined entirely in terms of the raw sensors available to the robot.

Specifically, our task-setting is defined as follows:
\begin{itemize}
    \item \textit{Rewards/Goal}: Because we do not assume access to any object tracker, the task goal must be defined solely using robot sensors.  Notably, for deformable plugs this eliminates the standard method of determining success via the robot's kinematics, since flexion is not observed by joint encoders.

    \item \textit{Gripper Pose}: Episodes begin with the plug in-hand, and roughly $5$ cm from the socket opening.  This work therefore assumes the plug can be reliably grasped using existing methods, but this is an important area of future work (\sref{sec:discussion}).
    \item \textit{Socket Pose}: Episodes have $2-4$ cm variability in the starting pose of the socket.
\end{itemize}

From anecdotal evidence we have found that industrial engineers often dismiss using robots for assembly tasks that ``require too much finesse''.
We believe learning methods are capable of advancing this frontier, and have designed two tasks within the setting described above that emphasize \textit{multi-modal} control in different ways:
\begin{enumerate}
    \item A \textit{peg insertion} task inspired by the \href{https://www.siemens.com/us/en/home/company/fairs-events/robot-learning.html}{Siemens challenge}\footnote{https://www.siemens.com/us/en/home/company/fairs-events/robot-learning.html}, which involves inserting a round peg into a round hole with $0.5$ mm clearance.  This task emphasizes force-sensitive probing behavior and accurate visual alignment.
    \item A \textit{clip insertion} task, which requires inserting a deformable clip with two prongs into square holes of a housing.  The robot must simultaneously pry and visually align multiple object parts whose pose cannot be inferred deterministically from the gripper pose.
\end{enumerate}

\section{RELATED WORK}
\label{sec:related_work}
Insertion is a well studied problem in both the machine learning and control communities.
Control approaches typically follow a similar format \cite{kronander2014task,johannsmeier2018framework,zhang2017peg}:
Decompose the task into discrete phases for approach, search, align, and insert, and design a state-machine and a collection of Cartesian impedance controllers to track the high-level and low-level behaviors, respectively.
The main challenge faced by these approaches in our setting is that the individual search and align behaviors must be made robust to uncertainty in the socket pose, which can be exacerbated by the lack of precision in the forward kinematics relative to the peg-clearance.
This can cause the plug to bind during insertion in narrow-clearance problems, motivating various wiggling and search behaviors \cite{johannsmeier2018framework}.
\cite{johannsmeier2018framework} addresses these issues with adaptive impedance control for insertion, and parameterized oscillations for wiggling the plug into the socket, and \cite{kronander2014task} attempts to learn recovery behavior from human interventions using a Gaussian-Mixture-Model.
Neither used visual information, but this would typically be obtained using an external object tracking system if required.
This family of approaches still require a controls-engineer to design the full insertion behavior (and tracking system), and uses learning only for parameter adaptation.

Full machine-learning approaches vary in state representation, but generally eschew manual skill-decomposition in favor of a global policy.
\cite{gullapalli1992learning,gullapalli1994learning} were the first to employ RL for learned compliance control on the peg-in-hole problem.
\cite{gullapalli1994learning} used a REINFORCE-like algorithm to train a stochastic neural-network which directly outputs position references for a low-level admittance controller.
Our approach shares several aspects with this pioneering work, including the use of a neural network policy which outputs continuous actions and the use of haptic feedback.
However, our approach adds modern (convolutional) deep-networks and uses a value function to reduce the variance of policy-gradient estimates, which together with the improvements described in \sref{sec:approach} significantly accelerate training.  Our RL agent also directly uses joint velocity actions, which relieves the need to write a Cartesian admittance controller, and is compliant to contact anywhere on the arm.

An alternative approach to enabling deep-RL on robots is Sim2Real: learning in simulation and transferring to the real system~\cite{rusu2016sim,bousmalis2017unsupervised}.
Sim2Real has shown promise for some tasks, e.g. grasp~\cite{rusu2016sim} and folding~\cite{matas2018sim}.
However both \cite{rusu2016sim} and \cite{matas2018sim} required considerable engineering effort to tune the simulator to match the physical system, and \cite{matas2018sim} was forced to side-step grasp using fake anchors between the cloth and the gripper.
Like \cite{matas2018sim} we are interested in complex manipulations of a grasped object, and struggled to produce stable simulations of grasp dynamics and force-feedback.\footnote{We used \href{http://www.mujoco.org/}{MuJoCo} (http://www.mujoco.org/) for physics simulation.}
In addition, the reliance on hand-crafted simulation violates our task setting, which aims to be end-user-programmable.

Guided Policy Search (GPS) \cite{levine2016end} can be used to train neural-network policies from raw pixels that solve insertion tasks, given optimal policies from a trajectory optimizer.
GPS is a powerful algorithm, but critically it requires low-dimensional features and a smooth cost function for the trajectory optimization step.
Our task setting does not admit the use of either of these.
\cite{finn2016deep} relaxes these requirements by replacing the engineered feature pipeline using an auto-encoder, and defines the cost as $L2$ distance in the feature space from a goal image.
Insertion tasks were not demonstrated using this method, but this would be an interesting avenue for future work.
The only work of which we are aware that handles \textit{some} socket position uncertainty without reliance on external features is \cite{inoue2017deep}.
However, this method uses a discrete and hand-defined Cartesian action space, and lacks a vision component which limits its tolerance to 3-5 mm.   \cite{inoue2017deep} also computes the reward from the robot kinematics, which restricts its applicability for deformable objects.

IN contrast, EDRIAD works in a scenario where rewards are given by the environment used by the demonstrator.
DQfD~\cite{hester2017learning} take this approach to extend DQN~\cite{Mnih:2015} with demonstrations and pretraining, but it only applies to discrete action domains. DDPGfD~\cite{vecerik2017leveraging} takes a similar approach for continuous action domains, but unlike our work, does not include pretraining, is not robust to environment stochasticity, and is not as sample-efficient.

\section{APPROACH}
\label{sec:approach}
Our approach is based on the DDPG algorithm \cite{lillicrap2015continuous}, which is an off-policy actor-critic algorithm which uses neural-networks to parameterize both the actor and critic, and uses the action-gradient $\frac{\partial Q}{\partial a}$ of the critic to train the actor.
Without changing this core mechanism, we introduce several modifications in order to satisfy our task setting and to maximize sample-efficiency:

\begin{enumerate}[label=\arabic*)]
    \item Train a classifier to compute task rewards from images, which is re-used to extract feature information.
    \item Add positive \textit{and} negative demonstrations of the task being solved.
    \item Add a behaviour cloning loss on the actor.
    \item Add demonstration-classification and episode-progress losses on the critic.
    \item Replace standard temporal-difference (TD) loss with a distributional critic loss \cite{bellemare2017distributional}.
\end{enumerate}

\subsection{Visual Features and Rewards}
\label{sub:visual_features}
In simulated manipulation tasks, the reward is often computed as a function of poses.
In a dynamic environment or with deformable plugs, this approach would require a devoted tracking system for both socket position and plug deformation.
To avoid this requirement, we instead train a convolutional network to detect if the plug is inserted from images obtained from an in-hand camera.

In our experiments we found that it was possible to train accurate insertion-detectors using a surprisingly small number of \textit{sequential} images obtained during a few minutes of tele-operation.
In addition, we found that with several additional constraints, these detector features could be co-opted to act as visual features for \textit{control}, in addition to classification. This was somewhat surprising given that the feature-losses did not include any explicit notion of action, time, or position.  This differs from the conclusions of \cite{finn2016deep} and \cite{jonschkowski2017pves}, and we attribute our success to the presence of joint velocity in the observation, and to the \textit{last-inch} task setting.
Previous results on training convolutional agents from scratch purely using reinforcement learning has not yet proven sufficiently data efficient to run directly on the real system \cite{lillicrap2015continuous,matas2018sim}.
However, it is a well-known phenomenon that internal activations of a neural network can often be reused for another related task \cite{yosinski2014transferable}.

In our preliminary experiments at reusing classification features for control, we found the features to be highly sensitive to visual noise and minor variations in the agent's behavior.
To mitigate this effect we added three additional losses with the following intuitions:

\renewcommand{\labelenumi}{\alph{enumi})}
\begin{enumerate}
    \item The mean activation should be 0 so the features have similar magnitudes across training runs.
    \item The feature covariance should be identity, forcing the features to be uncorrelated, and therefore contain as much independent information as possible.
    \item The features should not change if we add noise to the image or change its color or brightness.
\end{enumerate}

Note that (a) \& (b) are analogous to the KL-divergence between the feature-distribution and a standard-Normal distribution, as in Variational Auto-Encoders \cite{kingma2013autoencoding}.
We implemented this loss using batch statistics because our representation layer is non-stochastic, and for consistency with our other regularizers.
Our approach is also empirical (operating over batch statistics) rather than analytic (per-item), which puts less pressure on the individual features to be unimodal.

Denoting the image as $x$, feature extraction network with parameters $\theta_v$ as $f_{\theta_v}(\mathbb{R}^{hwc}) \rightarrow \mathbb{R}^k$, the reward classifier network with parameters $\theta_r$ as $g_{\theta_r}(\mathbb{R}^k) \rightarrow [0,1]$, and the image noise addition operator as $\nu$, then the loss for training the convolutional-network is:
\begin{equation*}
\begin{split}
L =& L_{ce}(g_{\theta_r}(f_{\theta_v}(x)), \delta^{+}) \\
& + \lambda_{r} L_2  \\
& + \lambda_{\mu} |\mathbb{E}_{b} (f_{\theta_v}(x))|^2_2 \\
& + \lambda_{\Sigma} |\mathbb{E}_{b} (f_{\theta_v}(x) f_{\theta_v}(x)^T) - \mathbb{I} |^2_2 \\
& + \lambda_{\sigma} |f_{\theta_v}(x) - E(\nu(x))|^2_2,
\end{split}
\end{equation*}
where $L_{ce}$ denotes cross-entropy,  $L_2$ is a regularization loss on magnitude of all network weights, $\lambda$ are hyper-parameters tuning relative importance of losses, $\delta^{+}$ is a soft insertion target, and $\mathbb{E}_{b}$ is an expectation over a batch of training samples.
We used $\delta^{+}=0.999$ for the correct category and $0.001$ otherwise, $\lambda_{r} = 10^{-3}$, $\lambda_{\mu} = 10^{-2}$, and $\lambda_{\Sigma} = \lambda_{\sigma} = 1$.
The use of soft targets means that a perfectly trained network should output logits of approximately $\pm 3.8$ for inserted and non inserted states rather than $\pm \infty$, which has been shown to reduce over-fitting~\cite{aghajanyan2017softtarget}.

We included $3$ kinds of noise which operated on RGB encodings of our images scaled between $0$ and $1$.
To modify the overall luminance we apply a $\gamma$ transform on all channels where $\gamma=exp(\mathbb{N}(0, 0.5)$.
To change the color balance we apply a different $\gamma$ transform on each channel where $\gamma=exp(\mathbb{N}(0, 0.2)$.
Lastly we add an uncorrelated pixel noise to each channel from normal distribution $\mathbb{N}(0, 0.05)$.

To decrease a chance of false-positives during run-time the agent receives positive reward if the output probability is $88\%$ (logit $>2$).
In practice however, agents were occasionally able to exploit false positives of the reward function.
We assume this is because the agents sees the activations within the detector and therefore has perfect information to exploit it.
To remedy this issue further we trained two models $g_{\theta_r^1}(f_{\theta_v^1})$ and $g_{\theta_r^2}(f_{\theta_v^2})$, and used only the features from $f_{\theta_v^1}$ as observations.
The agent received positive rewards only when both models exceed $88\%$ probability, which prevented reward delusions and further reduced false-positives:
\begin{equation*}
    r_t := \left\{
        \begin{array}{cl}
            1, & \left(g_{\theta_r^1}(f_{\theta_v^1}(x_t)) > 0.88 \right) \, \land \,
                 \left(g_{\theta_r^2}(f_{\theta_v^2}(x_t)) > 0.88 \right) \\
            0, & \text{otherwise} \\
        \end{array}
    \label{eq:reward_func}
    \right.
\end{equation*}
In our experiments the reward detector and thus the visual features were trained ahead of time and were fixed during the RL phase.
This sped up the training significantly as the convolutional network had to be evaluated only \textit{once} for each transition to compute the features and reward (which are then stored in replay), instead of requiring full batch forward and backward passes for each learning update.\footnote{As detailed in \sref{sub:ddpg_details} we perform roughly $10000$ learning updates per transition ($40$ batches of size $256$), so this results in a considerable speedup.}

\subsection{Collecting Demonstrations}
\label{sub:collecting_demonstrations}
Since the reward detector is focused on goal classification, neither the visual features nor rewards capture a smooth representation of goal-distance.
This poses a challenge for the RL phase, since the agent has no strong exploration cue.

Without extra information our tasks would be very difficult due to the hard exploration problem of finding a socket with only a sparse reward signal.
Following \cite{hester2017learning}, we use human demonstrations to address this issue.
To gather demonstrations we use a 6D mouse (SpaceNavigator) and interpret the motion as Cartesian commands which we translate to joint actions using the known kinematics of the arm.
We provide $20-30$ trajectories of successful solutions of the task, and the same number of \textit{negative} trajectories, in which we do not solve the task and instead probe different possible failure modes, \eg inserting only one clip-prong.
In \sref{sec:experiments} we compare performance with and without these negative demonstrations.
This teaching process takes less than 30 minutes of wall-clock time.

\subsection{Demonstration Critic Losses}
\label{sub:demo_loss}
We add two critic losses to extract as much signal as possible from the demonstration data. The first loss is a demonstration-classification loss, which forces the network to accurately predict whether sampled demonstration transitions are from the positive or negative set:
\begin{equation*}
L_{c} = \lambda_{c} \delta^{e} L_{ce}(C^{c}(s_{t-1}, a_{t-1}), \delta^{c}),
\end{equation*}
where $C^{c}$ is a head of the critic network outputting demonstration-label probabilities, $\delta^{e}$ is a mask which is $1$ for expert transitions and $0$ otherwise,  $\delta^{c}$ denotes the binary label on those transitions.
We used $\lambda_{c}$ of $100$.

The second loss is an episode progress loss where the critic estimates how far into an episode a given transition occurred.
We assign the fraction through the episode, $\chi$, to every step in a successful demonstration.
The critic predicts this number and we add an L2 loss on the error:
\begin{equation*}
L_{p} = \lambda_{p} \delta^{c} |C^{p}(s_{t-1}, a_{t-1}) - \chi|_2^2, \\
\end{equation*}
where $C^{p}$ is a head of the critic network and $\lambda_{p} = 1000$.

\subsection{Distributional Q-function}
\label{sub:distributional_critic}
We used a distributional critic which predicted the probability distribution of Q-values over $60$ bins which were evenly spaced between $0$ and $1$.
The learning loss is then the KL-divergence between our current estimate and a projection of the 1-step return on the current bins \cite{bellemare2017distributional}.

The TD-update loss becomes:
\begin{equation*}
L_{TD} = KL \left (
    Q(s_{t-1}, a_{t-1}),
    \Phi \left( r_t + \gamma_t Q^{target}(s_{t}, \pi^{target}(s_{t}) \right )
\right )
\end{equation*}
Where $Q$ is the value function, $\pi$ is the policy, $s$ are the sampled states, $a$ is the sampled action, $r$ is the reward, $\gamma$ is the discount, $\Phi$ is an operator which projects a distribution on a our set of bins.
We used $\gamma$ of $0.95$.

\subsection{Behaviour cloning}
\label{sub:behaviour_cloning}
We add a loss to the actor network for the policy to imitate the actions from the positive expert demonstrations.
This loss
is especially important during pretraining and the beginning of learning (see Fig.~\ref{fig:ablations}).
\begin{equation*}
L_{BC} = \lambda_{BC} \delta^{e} \delta^{c} |\pi(s_{t-1}) - a_{t-1}|_2^2
\end{equation*}
We used $\lambda_{BC}$ of 5.

\subsection{Summary of losses}
Overall the losses for the critic and actor are:
\begin{equation*}
\begin{split}
L_{critic} &= L_{TD} + exp(-S / \lambda_{e}) (L_{c} + L_{p}) \\
L_{actor} &= max(S^2 / \lambda_{AC}, 1) L_{AC} + exp(-S / \lambda_{e}) L_{BC} \\
L_{AC} &= -Q(s_{t-1}, \pi{s_{t-1}}),
\end{split}
\end{equation*}
where $L_{AC}$ is the actor-critic loss, $S$ is the number of steps in the environment, and $\delta^{e}$ is a mask which is $1$ for expert transitions and $0$ otherwise.
Both losses act on a schedule which decays the effect of the demonstration based losses over time.
We used $\lambda_{e}$ and $\lambda_{AC}$ of 500.

\subsection{DDPG Details}
\label{sub:ddpg_details}

As observation inputs we use positions, velocities and torques for each joint ($3 \cdot 7$ dimensions) as well as end-effector position  ($3$ dimensions) and $8$ visual features from the pretrained model.
The end-effector position was computed based on known kinematics.
These are concatenated into a 32 dimensional vector.
To improve learning efficiency, we offset the joint and effector positions so they would have 0 mean. We also scaled each input type to have similar scale across the input vector.

The agent uses a replay buffer which stores 1-step transitions and does uniform sampling.
Before the agent starts interacting with the environment we load all of the demonstrations in the replay buffer and perform pretraining~\cite{hester2017learning}. We never discard samples from the replay buffer.

At the beginning of training the supervised losses are often sufficient to train a successful policy.
However, the critic does not yet have a good gradient with respect to actions and applying it is detrimental to the policy.
Therefore, we apply a scaled down version of this gradient at the beginning of training and increase it over time.

\section{EXPERIMENTS}
\label{sec:experiments}

Our experiments address the following questions:
\begin{enumerate}[label=\arabic*)]
    \item How do the visual features affect performance on tasks with target position variability?
    \item How does our approach compare to simply performing behaviour cloning?
    \item What are the individual and aggregate effects of our auxiliary losses?
    \item How effective is pretraining and what makes it work?
\end{enumerate}

\subsection{Experimental setup}

We randomized the socket using a pan-tilt unit which moved to a different position at the beginning of each episode.
This position was selected by sampling an orientation from a uniform distribution with ranges $[-0.2,0.2]$ and $[-0.05,0.05]$ rads for the pan and tilt joints, respectively.
This corresponds to 2cm-4cm movement of the opening in both tasks.
We also present \textit{less randomized} versions of these tasks.
For these we use ranges $[-0.1,0.1]$ and $[-0.05,0.05]$ rads for the pan and tilt joints, respectively.

The agent sends control signals at 5Hz and episodes are limited to 40 steps (8s).
Episodes are terminated earlier if the agent has successfully solved the task.
For safety reasons the episode also terminates if the agent has failed by moving too far away from the intended workspace, or rotating its end effector too much.
All of our experiments run for 6000 steps, which corresponds to 20 minutes of environment interaction time of about 1.5h-2.5h of wall-clock time.
Each episode is split into three phases: environment reset, acting, and learning.
The difference between interaction and wall-clock time comes mainly from episode resetting and learning.
The variations in wall-clock time come from varying episode-length and therefore varying number of episode resets.

Our agent uses joint-velocity control and its actions are limited to $\pm 0.1$ rad/s.
These actions are not directly applied to the real robot but passed through an admittance layer which runs at 100Hz.
The velocity command to the arm is
\begin{equation*}
    v = v_a - \sigma \tau,
\end{equation*}
where $v_a$ is the agent's requested velocity, $\sigma$ is a compliance parameter, and $\tau$ is the gravity-compensated torque on the arm joints.

To optimize hyperparameters of our agents we used a simulation using the MuJoCo engine~\cite{todorov2012mujoco}.
We ran each set of hyperparameters 8 times and measured total reward achieved within 6000 steps.
We used the average across these seeds as a target in a Gaussian process optimizer~\cite{vizier}.

To visualize the performance we look at two metrics.
We define success rate as the ratio of successful episodes over the last 500 steps.
We also look at per-step reward to capture how quickly the task is solved.
We repeat every experiment four times for peg insertion tasks and six times for clip tasks.
We plot the mean and its standard error across the trials.

\subsection{Visual Features}

\begin{figure}[t]
  \centering
  \includegraphics[width=0.9\columnwidth]{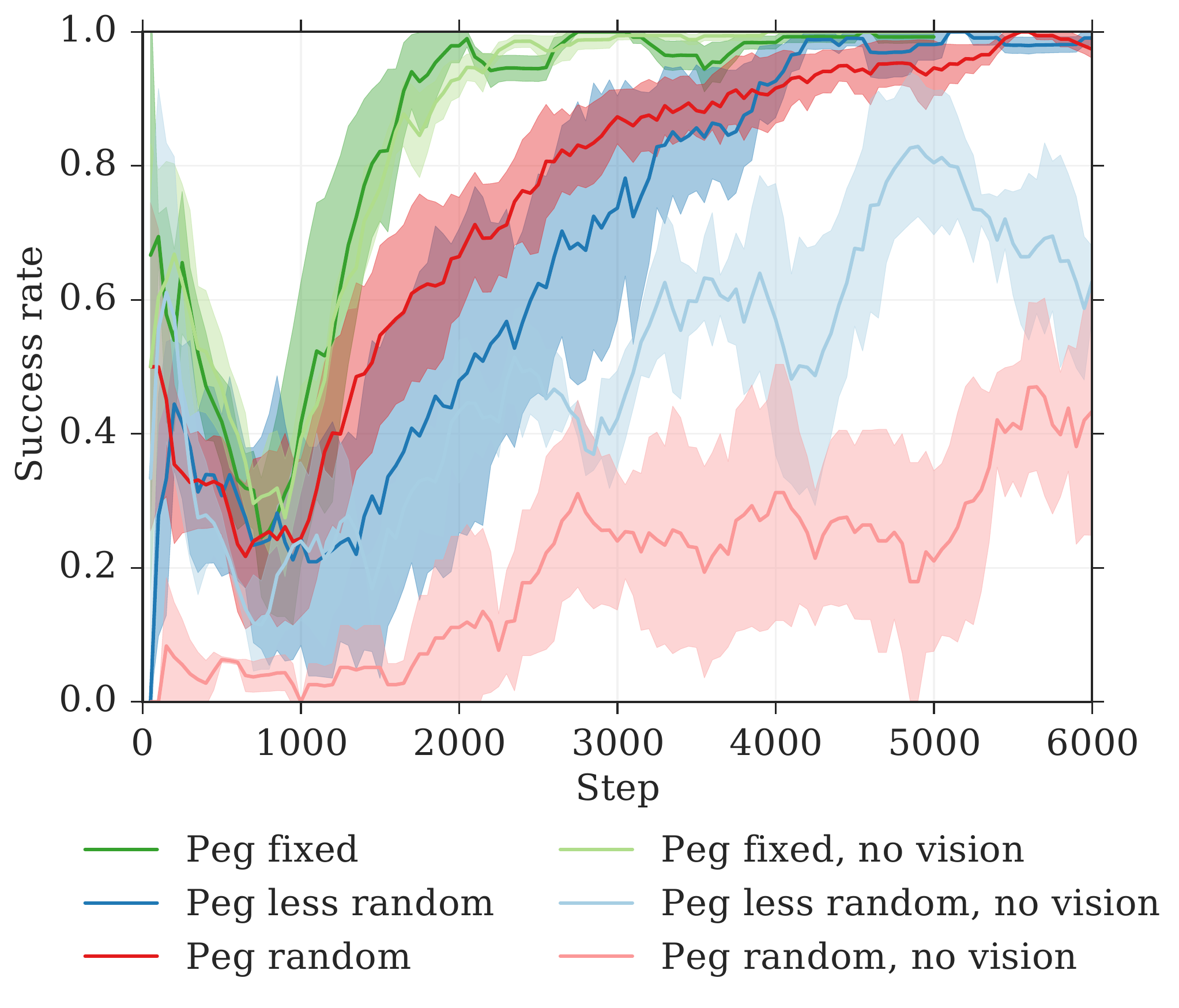}
  \includegraphics[width=0.9\columnwidth]{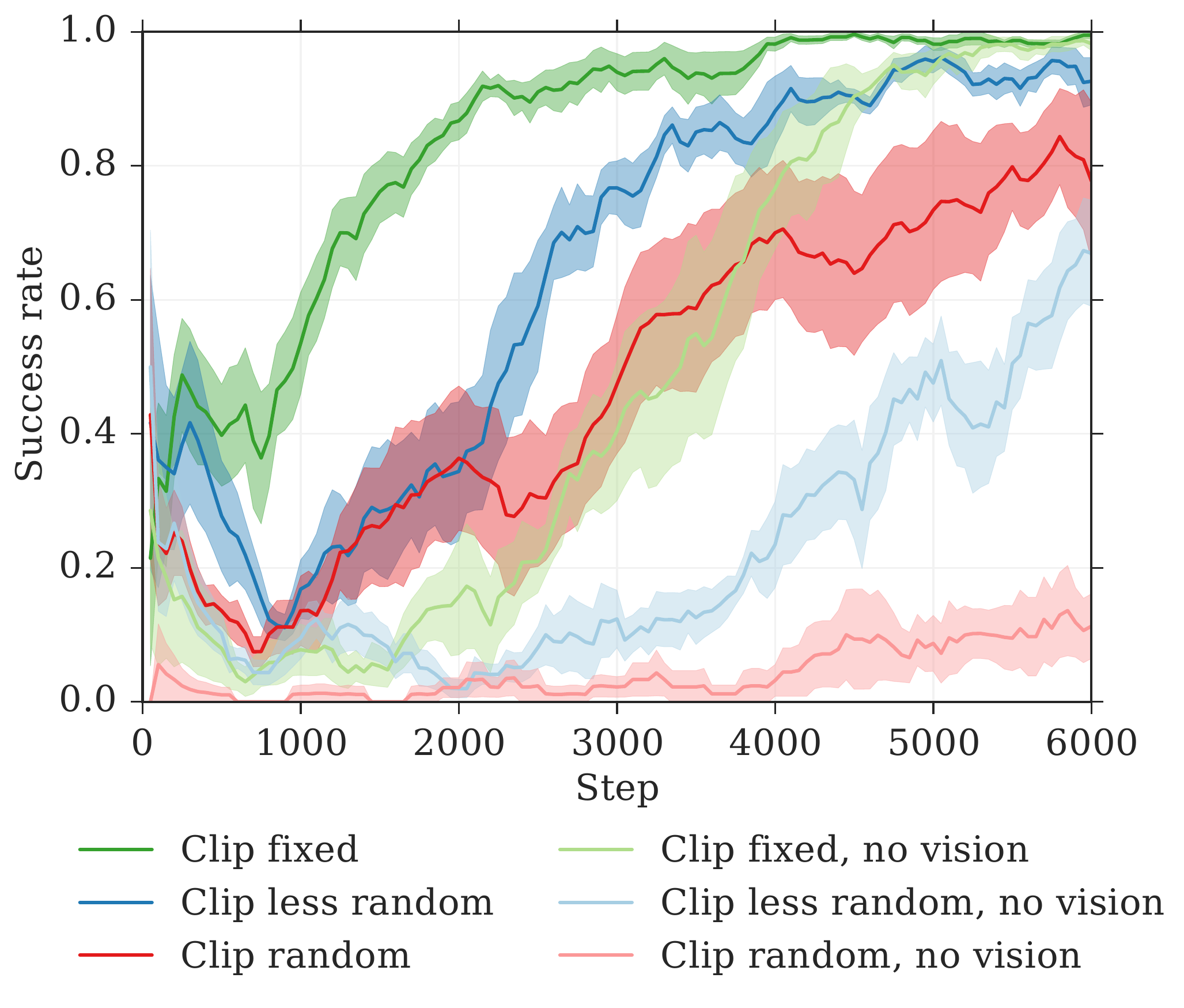}
  \caption{\small
  Learning curves show the success rate for the peg and clip task performed on the real robot.
  We show results for a fixed, less, and fully randomized socket as well as with and without visual features.
  The agent can solve the fixed position task without visual features, but they are necessary for solving the task with randomized socket position.
  }
  \label{fig:randomness}
\end{figure}

Figures~\ref{fig:randomness} shows that EDRIAD is able to learn both peg and clip insertion on the randomized version of the task.
On the peg insertion task our agent achieves average final success rate of $97\%$ and $77\%$ on the clip insertion task.
The clip task is more difficult due to the flexible nature of the clip, which makes the state of environment less observable.

The figure also shows that the randomized task cannot be solved by the agent without the visual features. They are necessary for the agent to
be robust to the variation in socket pose. When moving to the fixed location variant of the task, it can be solved both with and without visual features.

Additionally we investigate how the algorithm performs with a fixed goal and how the vision features affect the performance.
All agents were trained with the same hyper parameters as our simulation runs showed the same hyper parameters were optimal for both tasks and for agents without vision.
We do notice a small difference in optimal hyper parameters for tasks with a fixed goal.
This means the performance could be further improved, but we wanted to keep the comparison simpler across tasks.
For both tasks, Figure~\ref{fig:randomness} shows the best performance occurs in the case of fixed socket with vision features.
Both tasks can be learned within about 2000 steps which corresponds to about 6.5 minutes of interaction time.
The fixed socket version of the task can also be solved without the visual features.
As shown in Figure~\ref{fig:randomness}, visual features have a significant impact on agent performance, and the randomized task cannot be solved without them.

In addition, we found that agents without vision rely on probing behaviors to search for the opening(s).
In the clip case they these policies consistently attempt to first insert one prong and then the other.
While this is an interesting emergent behavior, it leads to very slow and unreliable policies on both tasks, compared to the agents with vision.

\subsection{Learning stability}

\begin{figure}[t]
  \centering
  \includegraphics[width=0.9\columnwidth]{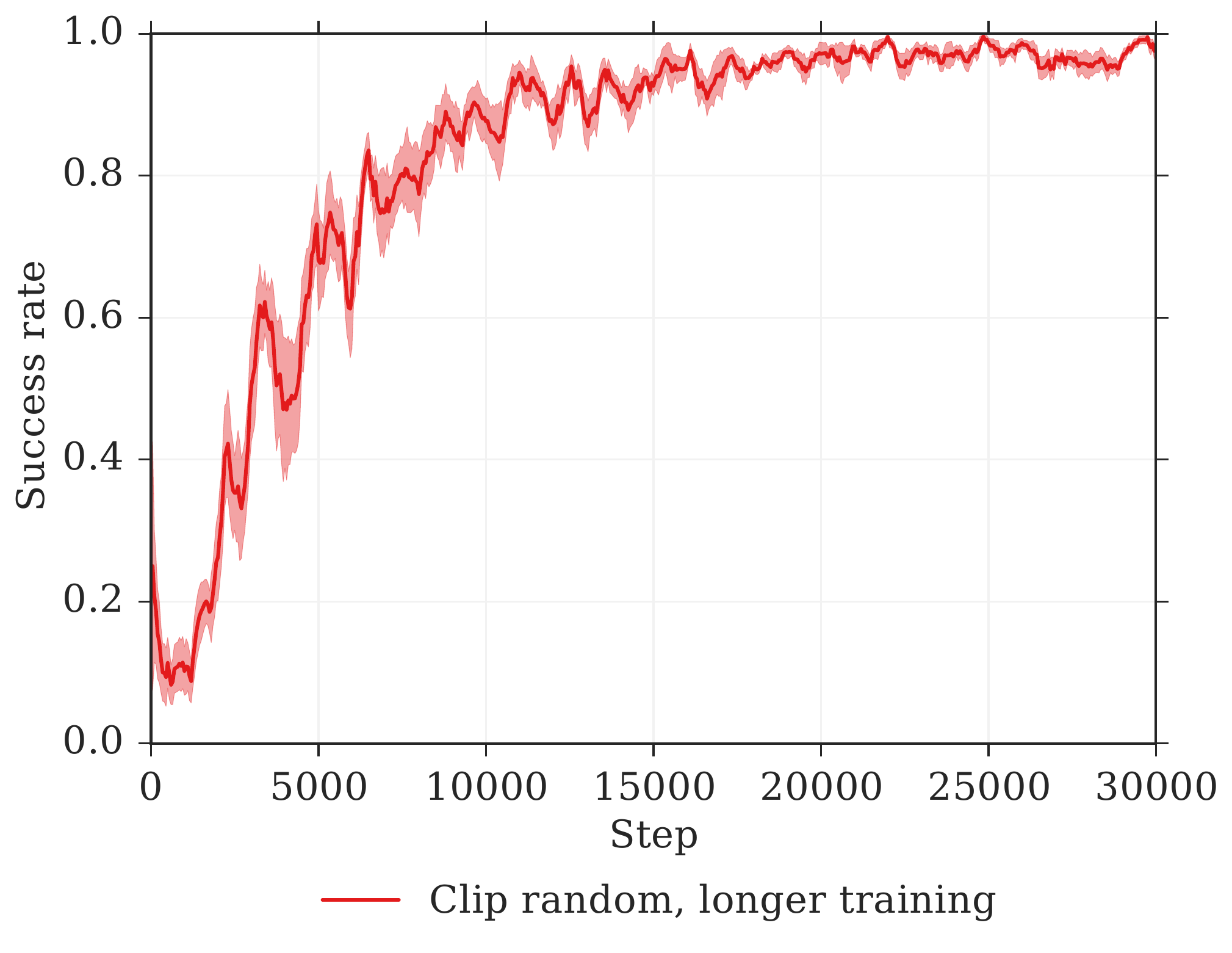}
  \caption{\small
  Success rate for the clip task over a longer time horizon for 8 runs.
  We see that the success rate keeps improving beyond 6000 steps in a stable way.
  }
  \label{fig:longrun}
\end{figure}

Figure~\ref{fig:randomness} shows that EDRIAD can learn a good policy within 6000 steps.
However it is not enough for it to fully converge.
To investigate the convergence of EDRIAD we let it train for 30000 steps (about 8 hours of wall-clock time).
Figure~\ref{fig:longrun} shows that EDRIAD keeps improving and reaches a success rate of $98\%$ at the end of the training.
We also note that the algorithm did not deteriorate or become unstable even though the hyperparameters were only tuned for 6000 step runs.

\subsection{Behaviour cloning}

A natural comparison to EDRIAD is to use behavior cloning, which can be implemented by a pure supervised loss on the actor.
To evaluate this approach, we trained the actor network three times and evaluated it at various points during the training.
Eventually the network over-fits, so we took the top performance from each run and report the mean and standard error of these best performances.

\begin{figure}[t]
  \centering
  \includegraphics[width=0.9\columnwidth]{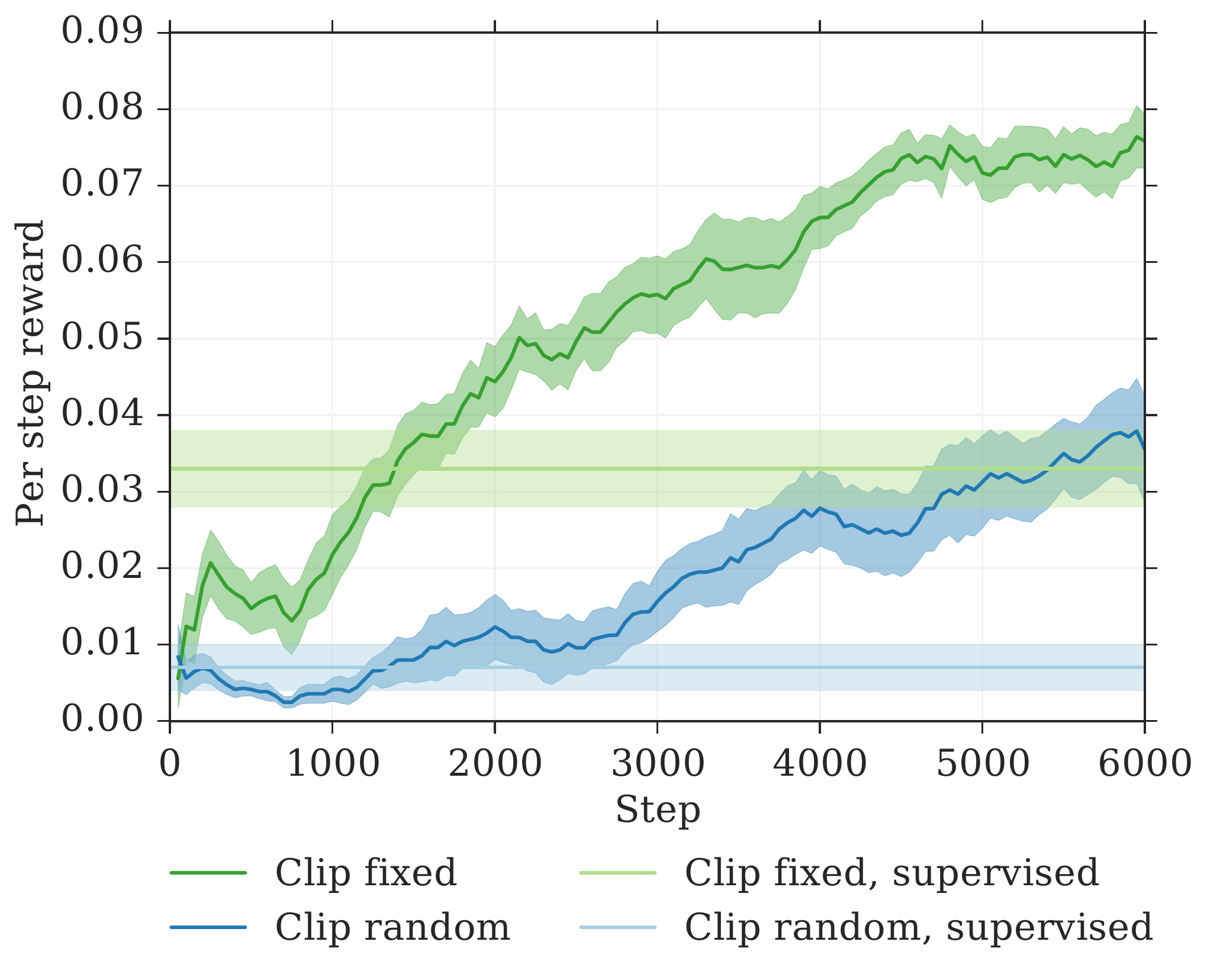}
  \caption{\small
  Learning curves show per-step reward for EDRIAD and behavioral cloning on the clip insertion task on the real robot.
  EDRIAD learns to insert the clip significantly faster than behavioral cloning in both the fixed and randomized socket variants of the task.
  }
  \label{fig:speed}
\end{figure}

The behavioral cloning agent using pre-trained vision features achieves a success rate of about $40\%$ on the randomized clip task and of about $80\%$ on the fixed socket version. 
However it can never improve beyond the demonstrations.
Figure~\ref{fig:speed} shows the per-step reward for both the fixed and randomized versions of the clip task.
In both cases our agent learns to exceed the performance of the supervised policy within 2000 steps, or 6 minutes of environment interaction.

In the fixed socket version of the task the agent learns to insert the clip significantly more quickly ($\sim3.2$s) than the demonstrator ($\sim7.3$s) or the behaviour cloning policy ($\sim7.6$s).
This demonstrates the main advantage of reinforcement-learning for these scenarios.

\subsection{Effects of auxiliary losses}

The primary novel components of EDRIAD vs. previous DPGfD work are the supervised losses on the critic and the actor, negative demonstrations and pre-training.
We demonstrate their effects on the real robot by measuring the performance without them on the randomized version of the tasks.
We also look at the individual contributions of each component in simulation, as it is not feasible to run all the experiments
required on the real system to evaluate each one individually.

\begin{figure}[t]
  \centering
  \includegraphics[width=0.9\columnwidth]{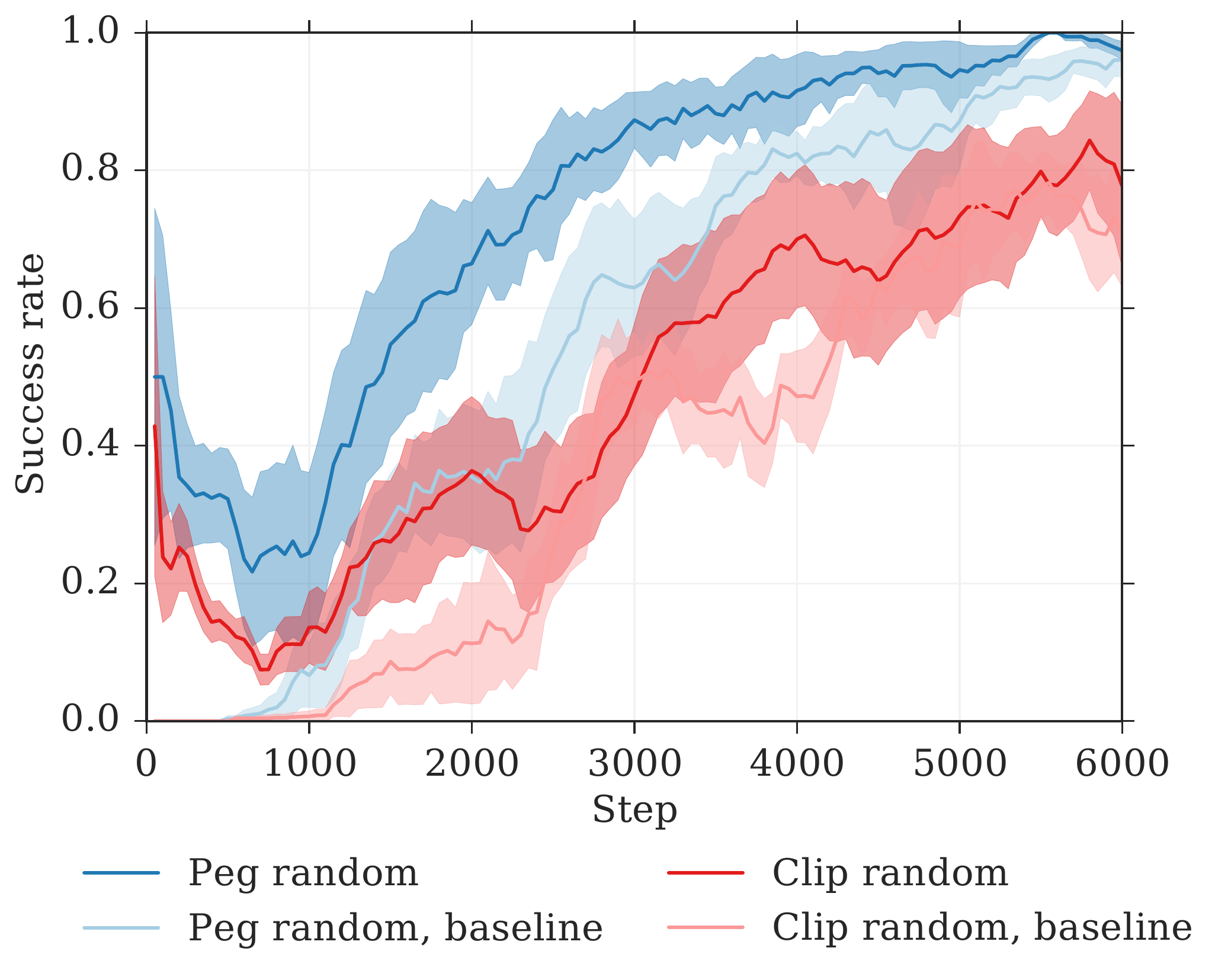}
  \caption{\small
  Learning curves for EDRIAD compared to a baseline on both insertion tasks on the real robot. The baseline does not have pre-training, supervised losses, negative examples in the replay buffer, or critic losses. These additional components enable EDRIAD to have higher performance on both tasks, particularly early in training.
  }
  \label{fig:baseline}
\end{figure}

Figure~\ref{fig:baseline} shows that removing all the contributions results in a large decrease in performance.
In particular, our modifications result in much better performance towards the beginning of training.
We do expect both algorithms to eventually converge to similar performances but the baseline algorithm reaches this performance more slowly.

\begin{figure}[t]
  \centering
  \includegraphics[width=0.9\columnwidth]{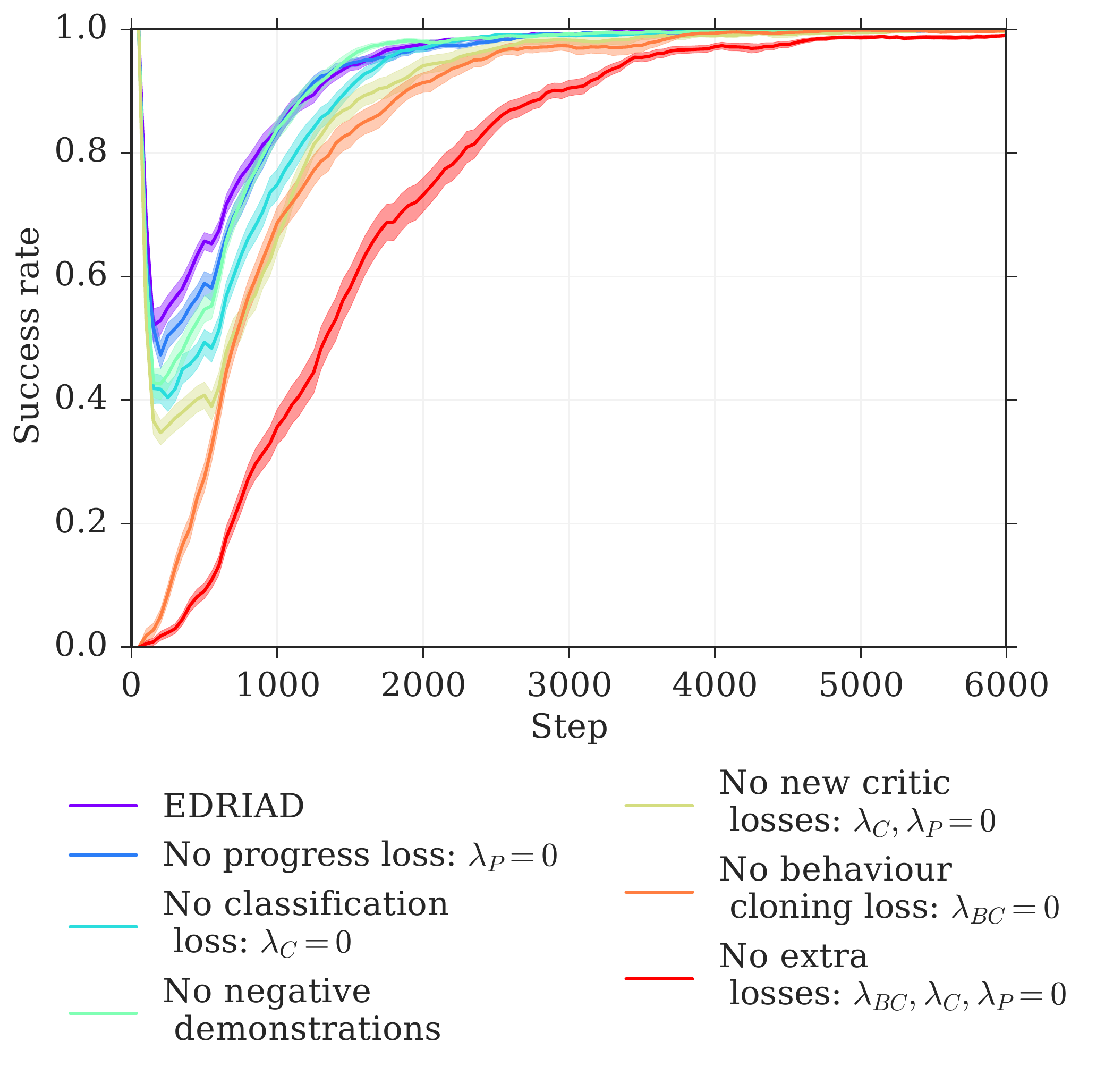}
  \caption{\small
    Ablated versions of EDRIAD on the randomized peg insertion task in simulation, averaged over 64 trials.
    Effect of individual losses are relatively minor, but the conjunction has a large effect.
  }
  \label{fig:ablations}
\end{figure}

We investigated the individual effects of each contribution on the simulated peg insertion environment with randomized socket position.
We ran 64 trials with different agent seeds and plot the mean and its standard error in Figure~\ref{fig:ablations}.
The episode progress and classification losses have small, but significant effects. When removing both losses, the effect is much bigger.
Similarly we see a negative effect of not using the negative demonstrations (and therefore the critic classification loss).
Removing the supervised loss has a large effect, even with all the other losses in place. This effect is largest in the early training immediately after pre-training, as the critic gradient needs sufficient number of samples to stabilize.
If we remove all of the extra losses, the performance suffers significantly and the training takes much longer to converge.

\subsection{Effectiveness of Pretraining}

\begin{figure}[t]
  \centering
  \includegraphics[width=0.9\columnwidth]{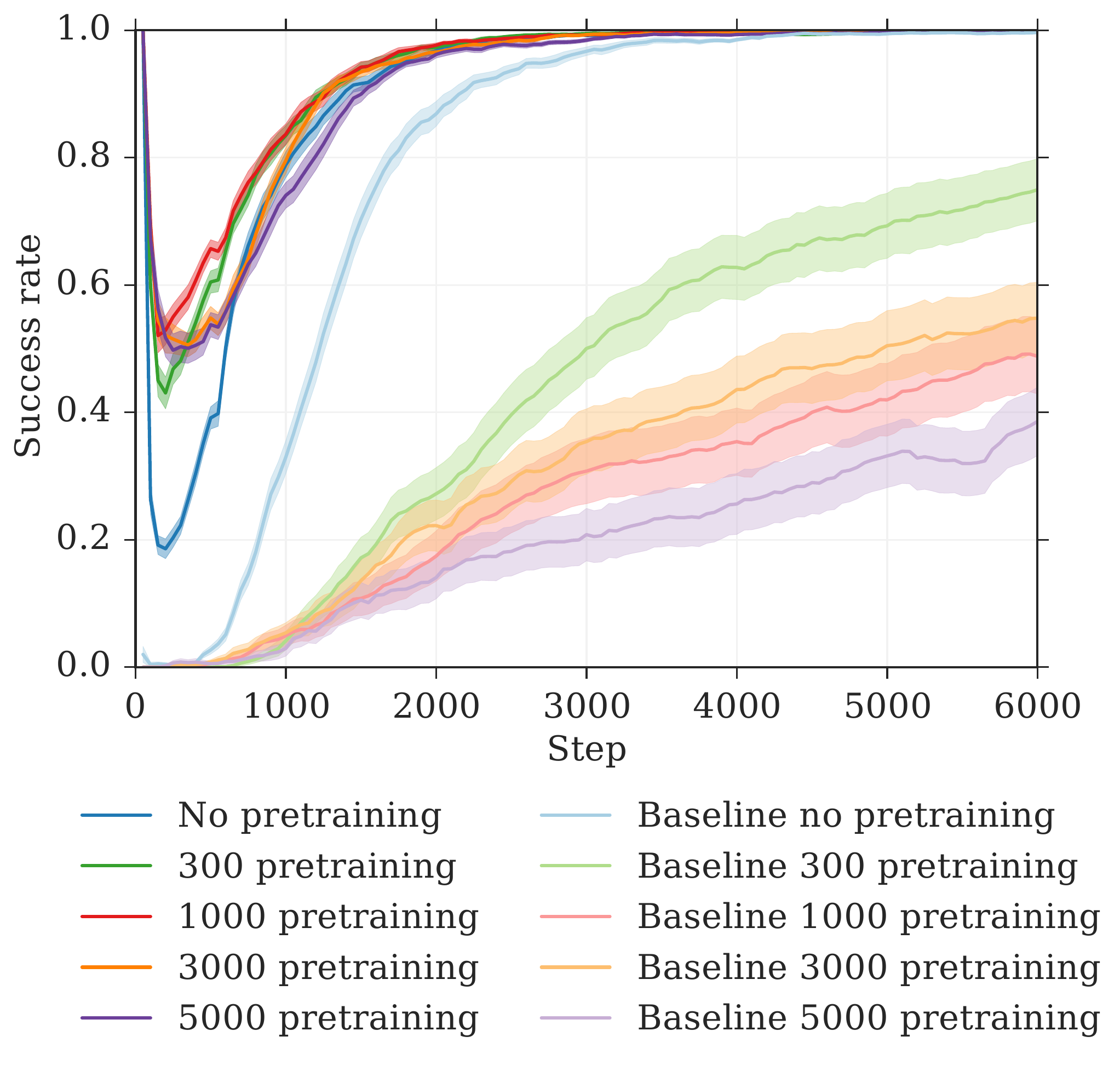}
  \caption{\small
  Learning curves for EDRIAD vs. baseline on the randomized clip insertion task in simulation, averaged over 64 trials.  Pretraining strictly hurts performance without stabilizing losses in \sref{sub:visual_features}, and is highly sensitive to the number of pretrain learning-steps.  With losses pretraining is consistently beneficial, and is robust to over-fitting.
  }
  \label{fig:pretraining}
\end{figure}

DDPG is an off-policy algorithm and therefore it is possible to train the agent from the demonstration data alone before interacting with the environment.
This pre-training is useful, but is potentially unstable without the right losses.
We compare EDRIAD to a baseline algorithm without the supervised losses on the actor and critic, negative demonstrations, or the gradual increase of the critic gradient into the actor.
We ran 64 trials on the simulated peg insertion task with randomized socket position (Fig.~\ref{fig:pretraining}).

With the full algorithm, there is a significant improvement when doing pre-training, even as it is scaled up to 1000 updates.
Beyond that, over-fitting starts to occur, but it does not catastrophically harm performance.
Without our additional losses, any pretraining harms the agent, and the harmful effects increase with more pretraining updates.

Finally, we tried using q-filtering \cite{nair2017overcoming} on behavior cloning, but we never observed any benefit.
We also tried training a vision model inside the agent, but found standard convolutional networks prohibitively slow to train, and small networks ineffective.

\section{DISCUSSION AND FUTURE WORK}
\label{sec:discussion}

The objective of this work is to solve insertion tasks in realistic industrial scenarios, and in a way that is entirely programmable by non-expert users.
Our results showed that our agent can reliably solve these tasks and is robust to variation in socket pose.
However, EDRIAD still required a carefully-engineered compliance controller, and considerable hyper-parameter tuning.
In addition, we assumed that the plug was already firmly grasped by the robot prior to starting the insertion phase, \eg from a feeder mechanism.
Both of these are restrictive assumptions that must be addressed for this work to reach a wider audience.

Anecdotally, we were surprised at the robustness of the final policies after RL training.
One failure mode for state-machine approaches is that unexpected perturbations can push the system into a situation in which the current controller's behavior is undefined, e.g. force-control in free-space.
While this can be handled by adding additional preemption logic to the state machine, implementation can be difficult and error-prone.
By contrast, EDRIAD (and end-to-end approaches in general) learns these behaviors smoothly and automatically.
In supplementary video at ({\scriptsize{\url{https://sites.google.com/view/dpgfd-insertion/home}}}) we show that our trained insertion policies are robust to variation in lighting, human perturbation of the robot, and even dynamic movement of the socket during an episode, despite not being trained for either of these cases.

Future work will explore simplification of the losses in \sref{sec:approach}, methods to reduce hyperparameter sensitivity and the need for parameter sweeps in simulation, and more principled methods of extracting visual features.

\section*{APPENDIX}
\label{sec:appendix}

\subsection{Network architectures}

\textit{Actor}:
Linear layers of 600, 400, 7 nodes with ReLu activations.
Output of the last layer are actions.

\textit{Critic}:
Linear layers of 800, 600, 60, 60 nodes with ReLu activations.
Output of the last layer is the soft max distribution over Q.

\textit{Reward network}:
Input image is RGB with size 128 by 128.
It first it is passed through a convolutional network with ReLu activations.
It has 3 layers with stride 3, kernel size 5 by 5 and number of channels is 16, 32, and 16 respectively.
After that the output is flattened and passed into a linear layer of size 8.
This layer is the feature layer which is later used in the agent.
After feature layer there is a single layer of size 32 with ReLu activations and last layer with a single node predicting the insertion.

\textit{Learning parameters}
Before interacting with the environment the agent performs $1000$ learning updates from the demonstration data to get a starting policy and critic.
Each learning update consists of sampling a mini-batch of $256$ from the replay buffer and performing a gradient descent step on it.
We used the ADAM optimizer with learning rates $10^{-5}$ for the actor and $0.0024$ for the critic.
We also $\beta_1$ of $0.88$ and $\beta_2$ of $0.92$.
During training we perform $40$ learning updates for each environment step \cite{popov2017data}.
However since these learning updates cannot be performed fast enough, they are always done after the end of episode.
Therefore each episode is split into three phases: environment reset, acting, and learning.
We maintain separate actor and critic target networks to stabilize learning and are updated every 10 learning updates \cite{lillicrap2015continuous}.

\clearpage
\printbibliography

\end{document}